%% file: 00_main.tex
\documentclass{article}
\usepackage{spconf,amsmath,graphicx}
\usepackage{multirow} 
\usepackage{makecell}
\usepackage{tikz}
\usepackage{arydshln}
\usepackage{enumitem}
\usepackage{hyperref}
\usepackage{amssymb}
\usepackage{soul}
\usepackage{caption}
\usepackage{svg}
\usepackage{orcidlink}

\title{Foundation Models Boost Low-Level Perceptual Similarity Metrics}

\name{Abhijay Ghildyal$^1$\orcidlink{0000-0003-1940-9626}, Nabajeet Barman$^2$\orcidlink{0000-0003-2587-7370}, Saman Zadtootaghaj$^2$\orcidlink{0000-0002-6028-8507}}

\address{
    $^1$Department of Computer Science, Portland State University, USA, abhijay@pdx.edu \\
    $^2$Sony Interactive Entertainment (PlayStation), \{Nabajeet.Barman,Saman.Zadtootaghaj\}@sony.com}

\begin{document}

\maketitle
\begin{abstract}
For full-reference image quality assessment (FR-IQA) using deep-learning approaches, the perceptual similarity score between a distorted image and a reference image is typically computed as a distance measure between features extracted from a pretrained CNN or more recently, a Transformer network. Often, these intermediate features require further fine-tuning or processing with additional neural network layers to align the final similarity scores with human judgments. So far, most IQA models based on foundation models have primarily relied on the final layer or the embedding for the quality score estimation. In contrast, this work explores the potential of utilizing the intermediate features of these foundation models, which have largely been unexplored so far in the design of low-level perceptual similarity metrics. We demonstrate that the intermediate features are comparatively more effective. Moreover, without requiring any training, these metrics can outperform both traditional and state-of-the-art learned metrics by utilizing distance measures between the features. \\ Code: \url{https://github.com/abhijay9/ZS-IQA}
\end{abstract}

\input{01_intro}

\input{02_related}

\input{03_experiments}

\input{04_conclusion}

\vfill\pagebreak

\bibliographystyle{IEEEbib}
\bibliography{00_main}

\end{document}

%% file: 01_intro.tex
\section{Introduction}
\label{sec:intro}

It is well established that neural networks trained on ImageNet~\cite{russakovsky2015imagenet}, using architectures like AlexNet~\cite{krizhevsky2012imagenet}, VGG~\cite{simonyan2014very}, EfficientNet~\cite{efficient}, and others, provide effective metrics for assessing low-level perceptual similarity~\cite{zhang2018perceptual,kumarbetter}. Recent advancements reveal that large vision models like CLIP~\cite{CLIP} and DINO~\cite{caron2021emerging}, originally designed for high-level tasks such as image recognition and multimodal understanding, are also effective in assessing mid to low-level perceptual similarity~\cite{wang2023exploring, fu2024dreamsim, croce2024adversarially}. For example, Fu \textit{et al.}~\cite{fu2024dreamsim} utilize the final embeddings of CLIP's~\cite{CLIP} ViT vision encoder and DINO to compute cosine distance, providing a similarity score between two images, and find application in tasks like image retrieval and synthesis. Other previous works \cite{croce2024adversarially, wang2023exploring}
also focus on the final embedding rather than the \textit{intermediate features}. The embedding is the final representation produced by the model, typically from the last layer, after processing the input image. These embeddings capture high-level semantic information and are used for tasks such as image-text matching, retrieval, and classification. Embeddings serve as global representations that summarize the entire input. In contrast, \textit{intermediate features} refer to activations within the model prior to the computation of the final embedding. These activations capture localized, lower-level information, including features like edges, textures, and various patterns. Consequently, for developing low-level perceptual similarity metrics, past approaches have consistently relied on the \textit{intermediate features} of models~\cite{zhang2018perceptual, Ding20, ghildyal2022stlpips}. Based on these observations, we pose the question: \textit{Could utilizing the \textit{intermediate features} of large foundation models result in the development of more effective low-level perceptual similarity metrics?} To address this question, this study explores several aspects, with the following key contributions:
\vspace{-0.05in}
\begin{enumerate}[noitemsep]
    \item We compare whether embeddings or \textit{intermediate features} are more effective for developing a more accurate and robust low-level perceptual similarity metric.
    \item Through evaluations on various datasets and across different distribution and distance measures, we demonstrate that foundation models like DINO and CLIP variants yield more accurate and robust metrics.
\end{enumerate}

%% file: 02_related.tex
\section{Related Work}
\label{sec:related}

\input{tables/metrics_emb_vs_feats}

Traditional metrics rely on pixel-level comparisons. PSNR measures direct pixel value differences, while SSIM~\cite{wang2004image} and its variants such as MS-SSIM~\cite{wang2003multiscale}, and FSIMc~\cite{zhang2011fsim} leverage the understanding of HVS and measure differences in luminance, contrast, and structure across local and global image regions, enhancing robustness to misalignments. 

In contrast, deep learning-based metric such as LPIPS~\cite{zhang2018perceptual} uses \textit{intermediate features} extracted from pre-trained networks like AlexNet~\cite{krizhevsky2012imagenet} and VGG~\cite{simonyan2014very} to assess perceptual similarity, aligning more closely with human judgments. The performance is further improved by fine-tuning on datasets specifically tailored for human perceptual similarity.

The LPIPS metric processes the \textit{intermediate features} of the two images using 1x1 convolution layers to produce weighted feature maps, and subsequently calculates the $l_2$ distance between these maps. However, recent research shows that using the same CNN-based backbones, such as VGG and EfficientNet, and directly comparing the \textit{intermediate features} with distribution comparison measures like Wasserstein Distance (WSD), Jensen-Shannon Divergence (JSD), and Symmetric Kullback-Leibler Divergence (SKLD) can yield improved perceptual similarity scores~\cite{liao2024image}. These enhanced metrics, based on distribution measures, align with human perceptual similarity without any additional training.

Other metrics such as DISTS~\cite{Ding20} adopt a design similar to the LPIPS(VGG)~\cite{zhang2018perceptual} but use measures akin to SSIM rather than $l_2$ distance between feature maps, thereby improving their metric's robustness. They also incorporate $l_2$-pooling layers, which blur \textit{intermediate features} to enhance DISTS's robustness. Another recent work investigates aliasing in \textit{intermediate features} resulting from imperceptible geometric misalignments~\cite{ghildyal2022stlpips}. By examining various neural network components, including max-pool, stride, and others, they aligned their LPIPS-based metric's sensitivity with human perception of an imperceptible misalignment between images.

Recent metrics~\cite{fu2024dreamsim, wang2023exploring} that employ foundation models like CLIP~\cite{CLIP} and DINO~\cite{oquab2024dinov} as backbones have mainly focused on using embeddings, rather than \textit{intermediate features}. Although \cite{camara2024measuring} proposed utilizing \textit{intermediate features} for CLIP variants, their evaluation and analysis were restricted to the TID2013 dataset. In contrast, our study offers a more comprehensive analysis by examining performance across three different datasets, including the larger PIPAL dataset, which encompasses both synthetic and real algorithmic distortions. We also evaluate various DINO and CLIP backbone architectures, utilizing a range of distribution and distance measures, and compare our results with existing low-level FR-IQA methods to demonstrate that using \textit{intermediate features} in VLMs are also effective for improved robustness.

\input{tables/metrics_distribution_measures}

%% file: tables/metrics_emb_vs_feats.tex
\begin{figure*}
    \begin{minipage}[b]{.3\linewidth}
        \centering
        \includegraphics[width=0.9\linewidth]{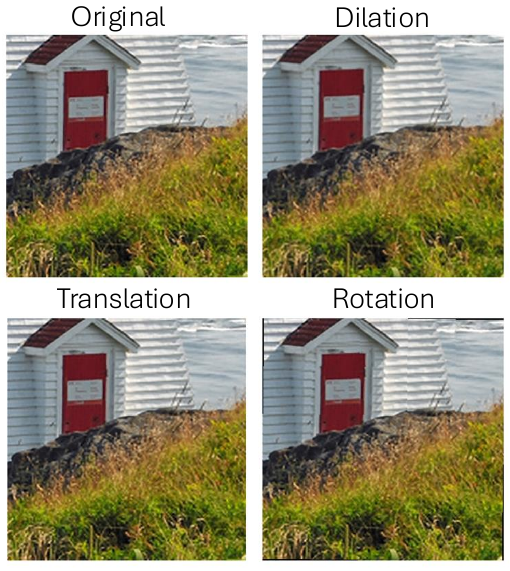}
        \vspace{-0.12in}
        \caption{Geometric transformations. Despite imperceptible changes, a metric's rank predictions often fluctuate.}
        \label{fig:geo_transforms}
    \end{minipage}
    \hfill
  \begin{minipage}[b]{.68\linewidth}
    \resizebox{0.98\textwidth}{!}{
        \begin{tabular}{@{}l@{\,}c@{\,}ccccc|cccc@{}}
        \hline
        
        \multirow{2}{9em}{Model} &\multicolumn{2}{c}{Type} &\multicolumn{4}{c|}{TID2013} &\multicolumn{4}{c}{PIPAL} \\
        \cline{2-11}
        &emb. &feats. &Ori. &Tra. &Dil. &Rot. &Ori. &Tra. &Dil. &Rot. \\
        \hline
        
        \multirow{2}{9em}{CLIP-RN50~\cite{CLIP}} &\checkmark & &0.518 &0.510 &0.499 &0.527 &0.320 &0.317 &0.315 &0.319 \\
        & &\checkmark &\underline{0.604} &\underline{0.616} &\underline{0.674} &\underline{0.638} &\underline{0.581} &\underline{0.538} &\underline{0.567} &\underline{0.570} \\
        
        \hdashline[1pt/5pt]
        
        \multirow{2}{9em}{CLIP-ConvNext~\cite{CLIP}} &\checkmark & &0.632 &0.625 &0.697 &0.654 &0.479 &0.477 &0.474 &0.482 \\
        & &\checkmark &\underline{0.658} &\underline{0.665} &\underline{0.753} &\underline{0.717} &\underline{0.598} &\color{blue}\underline{\textbf{0.552}} &\underline{0.587} &\underline{0.595} \\
    
        \hdashline[1pt/5pt]
        
        \multirow{2}{9em}{CLIP-ViT-B~\cite{CLIP}} &\checkmark & &0.693 &0.670 &0.748 &0.683 &0.531 &0.512 &0.526 &0.515 \\
        & &\checkmark &\color{blue}\underline{\textbf{0.758}} &\color{blue}\underline{\textbf{0.702}} &\color{red}\underline{\textbf{0.807}} &\color{blue}\underline{\textbf{0.784}} &\color{blue}\underline{\textbf{0.623}} &\underline{0.544} &\color{blue}\underline{\textbf{0.597}} &\color{blue}\underline{\textbf{0.593}} \\
    
        \hdashline[1pt/5pt]
        
        \multirow{2}{9em}{DINOv1-ViT-B~\cite{caron2021emerging}} &\checkmark & &0.783 &0.717 &0.792 &0.810 &0.622 &0.597 &0.615 &0.631 \\
        & &\checkmark &\color{red}\underline{\textbf{0.786}} &\color{red}\underline{\textbf{0.730}} &\color{blue}\underline{\textbf{0.804}} &\color{red}\underline{\textbf{0.814}} &\color{red}\underline{\textbf{0.637}} &\color{red}\underline{\textbf{0.607}} &\color{red}\underline{\textbf{0.629}} &\color{red}\underline{\textbf{0.643}} \\
        
        \hdashline[1pt/5pt]
        
        \multirow{2}{9em}{DINOv2-ViT-B~\cite{oquab2024dinov}} &\checkmark & &\underline{0.727} &0.672 &0.745 &0.747 &0.514 &0.484 &0.508 &0.523 \\
        & &\checkmark &0.722 &\underline{0.679} &\underline{0.763} &\underline{0.766} &\underline{0.573} &\underline{0.523} &\underline{0.558} &\underline{0.580} \\
        
        \hline
        \end{tabular}
    }
    
    \captionof{table}{Comparison of SRCC scores for embeddings (emb.) and intermediate features (feats.) using Cosine distance across the TID2013 and PIPAL datasets with various CLIP and DINO backbones. In each subgroup, \underline{best model results} are underlined, with overall best results highlighted in \textcolor{red}{red} and second-best in \textcolor{blue}{blue}.}
    \label{tab:metrics_emb_vs_inter}
    \end{minipage}
    
\vspace{-0.15in}
\end{figure*}

%% file: tables/metrics_distribution_measures.tex
\begin{table*}[t]
    \centering
    \setlength{\tabcolsep}{2pt}
    \resizebox{0.9\textwidth}{!}{%
    \begin{tabular}{@{}l c cc cc cc cc | cc cc cc cc@{}}
    \hline
     \multirow{3}{5em}{Model} &\multirow{3}{*}{Dist.} & \multicolumn{8}{c|}{LIVE} & \multicolumn{8}{c}{TID2013} \\
     \cline{3-18}
     & & \multicolumn{2}{c}{Original} & \multicolumn{2}{c}{Translation} & \multicolumn{2}{c}{Dilation} & \multicolumn{2}{c|}{Rotation} & \multicolumn{2}{c}{Original} & \multicolumn{2}{c}{Translation} & \multicolumn{2}{c}{Dilation} & \multicolumn{2}{c}{Rotation} \\
    
    & & PLCC & SRCC & PLCC & SRCC & PLCC & SRCC & PLCC & SRCC & PLCC & SRCC & PLCC & SRCC & PLCC & SRCC & PLCC & SRCC \\
    
    \hline
    
    \multicolumn{1}{c}{\multirow{2}{9em}{\textit{\large \underline{Distribution Measures}}}}  &&&&&&&&&&&& \\
    &&&&&&&&&&&& \\[-3pt]
    VGG~\cite{simonyan2014very} & \multirow{4}{3em}{\hfil SKLD\\~\cite{liao2024image}} & 0.914 & 0.936 & 0.456 & 0.429 & 0.714 & 0.697 & 0.646 & 0.619 & 0.751 & \underline{0.780}& 0.231 & 0.285 & 0.667 & 0.656 & 0.523 & 0.531 \\
    EfficientNet~\cite{efficient} &  & 0.800 & 0.896 & 0.808 & 0.816 & 0.869 & 0.867 & 0.868 & 0.862 & 0.577 & 0.647 & 0.692 & 0.652 & 0.718 & 0.683 & 0.733 & 0.689 \\
    CLIP-ViT-B~\cite{CLIP} &  & 0.938 & \color{blue}\underline{\textbf{0.965}} & \underline{0.895} & \underline{0.894} & \color{red}\underline{\textbf{0.936}} & \color{blue}\underline{\textbf{0.942}} & 0.937 & 0.940 & 0.730 & 0.765 & 0.768 & \underline{0.717} & \color{red}\underline{\textbf{0.843}} & \color{blue}\underline{\textbf{0.815}} & \color{red}\underline{\textbf{0.849}} & \color{blue}\underline{\textbf{0.819}} \\
    DINOv1~\cite{caron2021emerging} &  & \underline{0.949} & 0.964 & 0.893 & 0.883 & 0.920 & 0.914 & \color{red}\underline{\textbf{0.943}} & \underline{0.941} & \underline{0.770} & 0.768 & \underline{0.769} & 0.709 & 0.828 & 0.788 & \color{blue}\textbf{0.848} & 0.813 \\
    
    \hdashline[1pt/5pt]
    
    VGG~\cite{simonyan2014very} & \multirow{4}{3em}{\hfil JSD\\~\cite{liao2024image}}  & 0.910 & 0.928 & 0.436 & 0.405 & 0.690 & 0.667 & 0.620 & 0.587 & 0.752 & \underline{0.774} & 0.209 & 0.268 & 0.649 & 0.639 & 0.500 & 0.513 \\
    EfficientNet~\cite{efficient} &  & 0.863 & 0.905 & 0.834 & 0.839 & 0.875 & 0.870 & 0.875 & 0.867 & 0.642 & 0.649 & 0.719 & 0.669 & 0.733 & 0.689 & 0.744 & 0.694 \\
    CLIP-ViT-B~\cite{CLIP} &  & 0.852 & 0.959 & 0.869 & \underline{0.879} & \underline{0.925} & \underline{0.936} & 0.932 & 0.934 & 0.670 & 0.764 & 0.742 & 0.710 & 0.809 & \underline{0.806} & 0.838 & 0.809 \\
    DINOv1~\cite{caron2021emerging} &  & \color{blue}\underline{\textbf{0.951}} & \underline{0.964} & \underline{0.889} & \underline{0.879} & 0.917 & 0.911 & \color{blue}\underline{\textbf{0.941}} & \underline{0.940} & \underline{0.774} & 0.768 & \underline{0.765} & \underline{0.704} & \underline{0.827} & 0.785 & \underline{0.847} & \underline{0.812} \\
    
    \hdashline[1pt/5pt]
    
    VGG~\cite{simonyan2014very} &\multirow{4}{3em}{\hfil WSD\\~\cite{liao2024image}}  & 0.916 & 0.933 & 0.457 & 0.428 & 0.714 & 0.692 & 0.649 & 0.619 & 0.764 & \underline{0.780}& 0.257 & 0.323 & 0.693 & 0.680 & 0.552 & 0.560\\
    EfficientNet~\cite{efficient} &  & 0.837 & 0.897 & 0.780& 0.775 & 0.879 & 0.864 & 0.868 & 0.852 & 0.625 & 0.656 & 0.629 & 0.586 & 0.736 & 0.685 & 0.741 & 0.684 \\
    CLIP-ViT-B~\cite{CLIP} &  & 0.950 & 0.964 & \color{red}\underline{\textbf{0.903}} & \color{blue}\underline{\textbf{0.903}} & \color{blue}\underline{\textbf{0.935}} & \underline{0.941} & 0.936 & 0.942 & 0.735 & 0.744 & 0.747 & 0.702 & 0.823 & \underline{0.795} & 0.824 & 0.794 \\
    DINOv1~\cite{caron2021emerging} &  & \color{red}\underline{\textbf{0.954}} & \color{blue}\underline{\textbf{0.965}} & \color{blue}\textbf{0.899} & 0.888 & 0.926 & 0.920 & \color{red}\underline{\textbf{0.943}} & \color{blue}\underline{\textbf{0.943}} & \underline{0.774} & 0.765 & \underline{0.778} & \underline{0.716} & \underline{0.832} & 0.791 & \color{blue}\underline{\textbf{0.848}} & \underline{0.811} \\
    
    \hline
    
    \multicolumn{1}{c}{\multirow{2}{9em}{\textit{\large \underline{Other Distance Measures}}}}  &&&&&&&&&&&& \\
    &&&&&&&&&&&& \\[-3pt]
    CLIP-ViT-B~\cite{CLIP} &\multirow{2}{3em}{\hfil $l_2$}  & \underline{0.947} & \color{red}\underline{\textbf{0.968}} & \color{blue}\underline{\textbf{0.899}} & \color{red}\underline{\textbf{0.909}} & \underline{0.924} & \color{red}\underline{\textbf{0.947}} & \underline{0.919} & \color{red}\underline{\textbf{0.946}} & \color{red}\underline{\textbf{0.831}} & \color{red}\underline{\textbf{0.793}} & \color{blue}\underline{\textbf{0.797}} & \color{blue}\underline{\textbf{0.727}} & \color{blue}\underline{\textbf{0.842}} & \color{red}\underline{\textbf{0.821}} & \underline{0.844} & \color{red}\underline{\textbf{0.824}} \\
    DINOv1~\cite{caron2021emerging} &  & 0.937 & 0.964 & 0.896 & 0.891 & 0.906 & 0.914 & 0.915 & 0.940 & \color{blue}\textbf{0.824} & \color{blue}\textbf{0.792} & 0.794 & 0.720 & 0.821 & 0.790 & 0.834 & 0.813 \\
    
    \hdashline[1pt/5pt]
    
    CLIP-ViT-B~\cite{CLIP} &\multirow{2}{3em}{\hfil Cos.}  & \underline{0.937} & \color{blue}\underline{\textbf{0.965}} & 0.869 & 0.871 & 0.908 & 0.936 & 0.899 & 0.924 & 0.817 & 0.758 & 0.778 & 0.702 & \underline{0.840} & \underline{0.807} & \underline{0.833} & 0.784 \\
    DINOv1~\cite{caron2021emerging} &  & 0.933 & \underline{0.965} & \underline{0.896} & \underline{0.898} & \underline{0.911} & \underline{0.941} & \underline{0.911} & \color{blue}\underline{\textbf{0.943}} & \underline{0.820} & \underline{0.786} & \color{red}\underline{\textbf{0.800}} & \color{red}\underline{\textbf{0.730}} & 0.824 & 0.804 & \underline{0.833} & \underline{0.814} \\
    \hline
    \end{tabular}}
    \vspace{-0.05in}
    \caption{Performance comparison of different backbone architectures on LIVE and TID2013 datasets. Within each subgroup, \underline{better model results} are underlined, with the overall best performing model highlighted in \textcolor{red}{red} and the second-best in \textcolor{blue}{blue}.}
    \label{tab:metricsdistributionmeasures}
\vspace{-0.15in}
\end{table*}

%% file: 03_experiments.tex
\section{Experiments and Results}
\label{sec:experiments}

In our study, we evaluate the accuracy of several low-level perceptual similarity metrics by comparing their performance using correlation measures like PLCC, SRCC, and KRCC. Our experiments primarily focus on two key aspects: demonstrating that \textit{intermediate features} outperform embeddings for the low-level perceptual similarity task (Table~\ref{tab:metrics_emb_vs_inter}) and showing that CLIP and DINO models serve as superior backbones, generating more robust \textit{intermediate features} for enhanced low-level perceptual similarity metrics (Table~\ref{tab:metricsdistributionmeasures} and Table~\ref{tab:metrics_results_pipal}).

For the CLIP and DINO metrics with ViT backbones, we evaluate results using a square sliding window of size $224$ with a stride of $200$. For comparisons against previous state-of-the-art training-free metrics using VGG and EfficientNet, we use the authors' code~\cite{liao2024image} and perform evaluations without downsampling the input image (due to lack of information on the exact downsampling factor used by the authors for each dataset). Therefore, for uniformity, all our experiments are conducted on full-sized images without downsampling.

To investigate the robustness of these models, we adopt the experimental design from \cite{Ding20}, with the modification of applying geometric transformations to the distorted image while keeping the reference image unchanged. This approach better simulates real-world scenarios where distortions are more likely to be present in the image being analyzed. Specifically, we shift the distorted image horizontally to the right by 1\% of the pixels for translation, increase its scale by 1\% for dilation, and rotate it clockwise by 1 degree for rotation. A sample of this process is illustrated in Figure~\ref{fig:geo_transforms}.

\noindent \textbf{Embeddings vs. Intermediate Features.} Table~\ref{tab:metrics_emb_vs_inter} demonstrates that for all models, using \textit{intermediate features} (feats.) consistently results in better performance than using embeddings (emb.), suggesting that \textit{intermediate features} capture more relevant information for this task. This improvement is especially significant for DINOv1. Among the CLIP models, CLIP-ViT-B with using \textit{intermediate features} shows the best performance. DINOv1-ViT-B stands out as the best overall performer across both datasets and geometric distortion types, particularly when using \textit{intermediate features}. This indicates that DINOv1-ViT-B is highly effective for the low-level perceptual similarity task. Moreover, the results from Table~\ref{tab:metrics_emb_vs_inter} align with those in Table~\ref{tab:metrics_results_pipal}, where using \textit{intermediate features}, compared to embeddings (emb.), results in better correlation scores for both CLIP-ViT-B and DINOv1\footnote{In the rest of the paper, \textit{intermediate features} (feats.) are used, especially in Tables~\ref{tab:metricsdistributionmeasures} and \ref{tab:metrics_results_pipal}, unless the embedding (emb.) is specifically indicated.}.

\noindent \textbf{Comparison across different backbones.} As demonstrated in Table~\ref{tab:metricsdistributionmeasures} and Table~\ref{tab:metrics_results_pipal}, the ViT-B backbone models DINOv1 and CLIP-ViT-B generally perform the best across all three datasets, LIVE, TID2013, and PIPAL, and all geometric distortions, indicating higher consistency with human perceptual judgments. In contrast, VGG shows the lowest performance, particularly in robustness under geometric transformations.

\noindent \textbf{Comparison across distribution and distance measures.} Based on the results in Table~\ref{tab:metricsdistributionmeasures} and Table~\ref{tab:metrics_results_pipal}, the different distribution measures, i.e., SKLD, JSD, and WSD, do not notably affect the relative performance rankings of the models. Among the training-free models, the DINOv1 and CLIP-ViT-B backbones outperform the VGG and Efficient backbones, which had previously achieved state-of-the-art results using different distribution measures in a recent study~\cite{liao2024image}. The result is consistent across all three datasets, i.e., LIVE, TID2013, and PIPAL, demonstrating that the DINOv1 and CLIP-ViT-B features are both more accurate and robust. 

We also explore two other distance measures: $l_2$, and cosine distance (Cos.). Since, SKLD, JSD, and WSD include a weighted Euclidean norm, we evaluate the use of $l_2$ alone. As observed in Table~\ref{tab:metricsdistributionmeasures} and Table~\ref{tab:metrics_results_pipal}, the results using different distribution measures are almost the same as those obtained with $l_2$, suggesting that the distribution measures are dominated by the added Euclidean norm and that the adaptive weighting strategy requires further refinement. Additionally, previous studies show that cosine distance between embeddings aligns well with perceptual similarity scores, so we compute it between \textit{intermediate features} as well.

In summary, although CLIP-ViT-B performs better than DINOv1 on the LIVE and TID datasets with $l_2$ distance, DINOv1 exhibits significantly better performance on the larger PIPAL dataset. For cosine distance, SKLD, JSD, and WSD measures, DINOv1 consistently outperforms CLIP-ViT-B across all datasets. \textit{Thus, we recommend DINOv1 for low-level perceptual similarity tasks.}

\input{tables/metrics_results_pipal}

\noindent \textbf{Comparison against existing metrics.} Since none of these metrics were trained on the PIPAL Training dataset~\cite{pipal}, which includes a wide range of synthetic and computer vision algorithm-based distortions, it serves as a good test set for comparing all methods. Based on the results presented in Table~\ref{tab:metrics_results_pipal}, we can conclude that FSIMc is the best performing metric among the considered traditional methods. For the learned methods, we chose the most robust and high-performing metrics. Among the learned models we evaluated, ST-LPIPS(AlexNet) demonstrated superior performance. This might be due to the fact that LPIPS and ST-LPIPS are trained on the BAPPS dataset~\cite{zhang2018perceptual}, which includes a range of distortions from computer vision algorithms like super-resolution, frame interpolation, deblurring, and colorization. In contrast, DISTS is trained on the KADID-10k dataset~\cite{kadid10k}, which mainly includes synthetic distortions. 

Despite not being fine-tuned for low-level perceptual similarity tasks, training-free metrics with CLIP-ViT-B and DINOv1 backbones perform well. Among these, DINOv1 stands out as the top performer among all models, particularly in its handling of various geometric distortions.

\noindent \textbf{Comparison against ImageNet-ViT.} In terms of $l_2$ distance, CLIP-ViT and DINOv1-ViT outperform ImageNet-ViT~\cite{dosovitskiy2020vit} significantly. However, for cosine distance CLIP-ViT and ImageNet-ViT perform similarly while DINOv1 has a significantly better performance. Both DINOv1 and ImageNet-ViT are trained on the ImageNet dataset, highlighting the importance of this pre-training in producing better features. Furthermore, DINOv1 employs a self-distillation training method, where the student model learns to align its image representations with those of the teacher network across various augmentations, without requiring labeled data. As a result, for downstream tasks, the pretrained features from DINOv1 have been shown to outperform those from ImageNet-ViT, which is trained with supervision on ImageNet. We observed a similar trend in our work for low-level perceptual similarity.

\noindent \textbf{Embeddings of Adversarially Robust-CLIP.} There is growing interest in using adversarially trained models for perceptual similarity~\cite{ghildyal2023attackPercepMetrics,ghazanfari2023rlpips}. A recent study found that cosine distance between Robust-CLIP embeddings outperforms that between CLIP embeddings.~\cite{croce2024adversarially}. Consequently, we compared the embeddings of R-CLIP$_\text{T}$-ViT and CLIP-ViT. The results presented in Table~\ref{tab:metrics_results_pipal} show that R-CLIP exhibits superior performance in terms of SRCC and KRCC, although it lags behind in terms of PLCC scores. The pretrained adversarially robust CLIP is less sensitive to translation distortions but not as effective against dilation and rotation. At present, it remains inconclusive whether adversarial training improves both accuracy and geometric robustness. In the future, we plan to further explore adversarially robust backbones for the low-level perceptual similarity task.

%% file: tables/metrics_results_pipal.tex
\begin{table*}[t]
    \centering
    \resizebox{0.9\textwidth}{!}{%
    \begin{tabular}{@{}l c ccc | ccc | ccc | ccc@{}}
    \hline
    \multirow{3}{5em}{Model} &\multirow{3}{*}{Dist.} & \multicolumn{12}{c}{PIPAL} \\
    \cline{3-14}
    & &\multicolumn{3}{c|}{Original} &\multicolumn{3}{c|}{Translation} &\multicolumn{3}{c|}{Dilation} &\multicolumn{3}{c}{Rotation} \\
    
    & &PLCC & SRCC & KRCC & PLCC & SRCC & KRCC & PLCC & SRCC & KRCC & PLCC & SRCC & KRCC \\
    
    \hline
    
    \multicolumn{1}{c}{\multirow{2}{9em}{\textit{\large \underline{Traditional}}}}  &&&&&&&&&&&& \\
    &&&&&&&&&&&& \\[-3pt]
    PSNR &\multirow{3}{*}{-} &0.403 &0.407 &0.276 &-0.065 &-0.079 &-0.052 &0.064 &0.038 &0.026 &-0.024 &-0.042 &-0.028 \\
    MS-SSIM~\cite{wang2003multiscale} & &0.562 &0.562 &0.397 &-0.003 &-0.012 &-0.008 &0.318 &0.227 &0.153 &0.186 &0.108 &0.072 \\
    FSIMc~\cite{zhang2011fsim} & &\underline{0.609} &\underline{0.589} &\underline{0.416} &\underline{0.209} &\underline{0.190} &\underline{0.127} &\underline{0.368} &\underline{0.328} &\underline{0.220} &\underline{0.292} &\underline{0.254} &\underline{0.170} \\
    
    \hline
    
    \multicolumn{1}{c}{\multirow{2}{9em}{\textit{\large \underline{Learned}}}}  &&&&&&&&&&&& \\
    &&&&&&&&&&&& \\[-3pt]
    DISTS~\cite{Ding20} &\multirow{5}{*}{-} &0.580 &0.579 &0.407 &0.560 &0.558 &0.390 &0.582 &0.575 &0.403 &0.546 &0.539 &0.375 \\
    LPIPS(AlexNet)~\cite{zhang2018perceptual} & &0.586 &0.588 &0.412 &0.567 &0.567 &0.394 &0.539 &0.532 &0.366 &0.574 &0.573 &0.398 \\
    LPIPS(VGG)~\cite{zhang2018perceptual} & &0.611 &0.588 &0.415 &0.548 &0.527 &0.366 &0.588 &0.561 &0.392 &0.584 &0.558 &0.390 \\
    ST-LPIPS(AlexNet)~\cite{ghildyal2022stlpips} & &\underline{0.628} &\underline{0.631 }&\underline{0.448} &\underline{0.625} &\color{red}\underline{\textbf{0.628}} &\color{red}\underline{\textbf{0.445}} &\underline{0.618} &\underline{0.614} &\underline{0.434} &\underline{0.627} &\underline{0.630} &\underline{0.447} \\
    ST-LPIPS(VGG)~\cite{ghildyal2022stlpips} & &0.578 &0.580 &0.406 &0.572 &0.573 &0.401 &0.557 &0.550 &0.382 &0.577 &0.580 &0.407 \\
    \hline
    \multicolumn{1}{c}{\multirow{2}{9em}{\textit{\large \underline{Training-free}}}}  &&&&&&&&&&&& \\
    &&&&&&&&&&&& \\[-3pt]
    VGG~\cite{liao2024image} &\multirow{4}{3em}{SKLD\\~\cite{liao2024image}} &0.536 &0.563 &0.392 &0.233 &0.228 &0.152 &0.401 &0.411 &0.277 &0.353 &0.350 &0.235 \\
    EfficientNet~\cite{liao2024image} & &0.504 &0.503 &0.354 &0.453 &0.418 &0.289 &0.481 &0.452 &0.314 &0.474 &0.439 &0.305 \\
    CLIP-ViT-B & &0.600 &0.616 &0.438 &0.564 &0.557 &0.388 &0.598 &0.597 &0.421 &0.607 &0.596 &0.420 \\
    DINOv1 & &\underline{0.627} &\color{blue}\underline{\textbf{0.639}} &\color{blue}\underline{\textbf{0.458}} &\underline{0.632} &\underline{0.615} &\underline{0.437} &\underline{0.643} &\color{blue}\underline{\textbf{0.633}} &\color{blue}\underline{\textbf{0.452}} &\underline{0.681} &\color{blue}\underline{\textbf{0.660}} &\color{red}\underline{\textbf{0.476}} \\
    \hdashline[1pt/5pt]
    
    VGG~\cite{liao2024image} &\multirow{4}{3em}{JSD\\~\cite{liao2024image}} &0.530 &0.552 &0.383 &0.214 &0.208 &0.139 &0.387 &0.391 &0.263 &0.335 &0.328 &0.220 \\
    EfficientNet~\cite{liao2024image} & &0.520 &0.506 &0.356 &0.464 &0.428 &0.296 &0.488 &0.454 &0.316 &0.482 &0.444 &0.308 \\
    CLIP-ViT-B & &0.494 &0.610 &0.444 &0.528 &0.545 &0.390 &0.549 &0.588 &0.426 &0.590 &0.584 &0.422 \\
    DINOv1 & &\underline{0.630} &\underline{0.638} &\underline{0.457} &\underline{0.633} &\underline{0.613} &\underline{0.436} &\underline{0.645} &\underline{0.632} &\underline{0.451} &\underline{0.682} &\color{blue}\underline{\textbf{0.660}} &\color{blue}\underline{\textbf{0.475}} \\
    \hdashline[1pt/5pt]
    
    VGG~\cite{liao2024image} &\multirow{4}{3em}{WSD\\~\cite{liao2024image}} &0.558 &0.582 &0.407 &0.311 &0.305 &0.205 &0.458 &0.465 &0.316 &0.421 &0.418 &0.282 \\
    EfficientNet~\cite{liao2024image} & &0.525 &0.503 &0.354 &0.405 &0.365 &0.251 &0.488 &0.443 &0.308 &0.473 &0.423 &0.293 \\
    CLIP-ViT-B & &0.614 &0.627 &0.446 &0.578 &0.571 &0.400 &0.609 &0.608 &0.430 &0.619 &0.607 &0.429 \\
    DINOv1 & &\underline{0.633} &\color{red}\underline{\textbf{0.641}} &\color{red}\underline{\textbf{0.460}} &\underline{0.637} &\color{blue}\underline{\textbf{0.618}} &\color{blue}\underline{\textbf{0.439}} &\underline{0.647} &\color{red}\underline{\textbf{0.636}} &\color{red}\underline{\textbf{0.454}} &\color{blue}\underline{\textbf{0.683}} &\color{red}\underline{\textbf{0.661}} &\color{red}\underline{\textbf{0.476}} \\

    \hdashline
    ImageNet-ViT~\cite{dosovitskiy2020vit} &\multirow{6}{*}{$l_2$} &0.469 &0.444 &0.306 &0.239 &0.212 &0.143 &0.371 &0.335 &0.228 &0.345 &0.308 &0.209 \\
    CLIP-RN50 & &0.567 &0.558 &0.391 &0.498 &0.488 &0.335 &0.550 &0.540 &0.375 &0.546 &0.537 &0.373 \\
    CLIP-ConvNext & &0.122 &0.137 &0.091 &0.109 &0.121 &0.081 &0.099 &0.109 &0.073 &0.111 &0.124 &0.082 \\
    CLIP-ViT-B & &0.649 &0.619 &0.440 &0.600 &0.563 &0.393 &0.634 &0.601 &0.424 &0.634 &0.602 &0.425 \\
    DINOv1 & &\color{red}\underline{\textbf{0.683}} &\color{blue}\underline{\textbf{0.639}} &\color{blue}\underline{\textbf{0.458}} &\color{red}\underline{\textbf{0.664}} &\underline{0.616} &\underline{0.438} &\color{red}\underline{\textbf{0.679}} &\color{blue}\underline{\textbf{0.633}} &\color{blue}\underline{\textbf{0.452}} &\color{red}\underline{\textbf{0.698}} &\color{blue}\underline{\textbf{0.660}} &\color{blue}\underline{\textbf{0.475}} \\
    DINOv2 & &0.446 &0.416 &0.286 &0.340 &0.300 &0.203 &0.403 &0.363 &0.247 &0.371 &0.336 &0.228 \\

    \hdashline[1pt/5pt]
    
    ImageNet-ViT~\cite{dosovitskiy2020vit} &\multirow{9}{*}{Cos.} &0.657 &0.616 &0.438 &0.612 &0.557 &0.391 &0.641 &0.592 &0.418 &0.645 &0.598 &0.423 \\
    CLIP-RN50 & &0.610 &0.598 &0.426 &0.564 &0.552 &0.387 &0.609 &0.587 &0.415 &0.618 &0.595 &0.422 \\
    CLIP-ConvNext & &0.603 &0.581 &0.413 &0.565 &0.538 &0.377 &0.598 &0.567 &0.401 &0.600 &0.570 &0.404 \\
    CLIP-ViT-B & &0.663 &0.623 &0.444 &0.590 &0.544 &0.379 &0.641 &0.597 &0.421 &0.638 &0.593 &0.418 \\
    CLIP-ViT-B emb. & &0.522 &0.531 &0.369 &0.508 &0.512 &0.354 &0.517 &0.526 &0.365 &0.509 &0.515 &0.357 \\
    R-CLIP$_\text{T}$~\cite{croce2024adversarially} ViT-B emb. & &0.385 &0.547 &0.383 &0.382 &0.528 &0.367 &0.386 &0.455 &0.309 &0.384 &0.479 &0.331 \\
    DINOv1 & &\color{blue}\underline{\textbf{0.677}} &\underline{0.637} &\underline{0.456} &\color{blue}\underline{\textbf{0.654}} &\underline{0.607} &\underline{0.431} &\color{blue}\underline{\textbf{0.671}} &\underline{0.629} &\underline{0.448} &\color{blue}\underline{\textbf{0.683}} &\underline{0.643} &\underline{0.460} \\
    DINOv1 emb. & &0.663 &0.622 &0.444 &0.644 &0.597 &0.423 &0.658 &0.615 &0.437 &0.672 &0.631 &0.451 \\
    DINOv2 & &0.619 &0.573 &0.406 &0.576 &0.523 &0.366 &0.608 &0.558 &0.393 &0.625 &0.580 &0.410 \\
    
    \hline
    \end{tabular}}
    \vspace{-0.05in}
    \caption{Performance comparison of different metrics on the PIPAL dataset. Within each subgroup, \underline{better model results} are underlined, with the overall best performing model highlighted in \textcolor{red}{red} and the second-best in \textcolor{blue}{blue}.}
    \label{tab:metrics_results_pipal}
\vspace{-0.15in}
\end{table*}

%% file: 04_conclusion.tex
\section{Conclusion}
\label{sec:conclusion}

In this study, we performed a comprehensive evaluation of different distance measures between \textit{intermediate features} of various foundation models as training-free metrics for low-level perceptual similarity estimation. Our findings highlight that metrics using \textit{intermediate features} outperform those using embeddings, with DINOv1 being the overall top performer. Future work will focus on fine-tuning these features on perceptual similarity datasets for improved results.